\documentclass{article}

\usepackage{arxiv}

\usepackage[utf8]{inputenc} 
\usepackage[T1]{fontenc}    
\usepackage{hyperref}       
\usepackage{url}            
\usepackage{booktabs}       
\usepackage{amsfonts}       
\usepackage{nicefrac}       
\usepackage{microtype}      
\usepackage{lipsum}

\usepackage{graphicx}
\usepackage{amsmath}
\usepackage[noline,ruled]{algorithm2e}
\usepackage{array}
\usepackage{multirow}
\makeatletter
\newcolumntype{"}{@{\hskip\tabcolsep\vrule width 1pt\hskip\tabcolsep}}
\makeatother
\usepackage{amssymb}
\usepackage{pifont}
\newcommand{\cmark}{\ding{52}}
\newcommand{\xmark}{\ding{55}}
\usepackage[figuresright]{rotating}

\title{Computational Models for Academic Performance Estimation}

\author{
  Vipul Bansal\thanks{equal contribution} \\
  Dept. of Mechanical and Industrial Engineering\\
  Indian Institute of Technology\\
  Roorkee, India \\
  \texttt{vbansal@me.iitr.ac.in} \\
  \And
  Himanshu Buckchash\textsuperscript{$\ast$} \\
  Dept. of Computer Science and Engineering\\
  Indian Institute of Technology\\
  Roorkee, India \\
  \texttt{hbuckchash@cs.iitr.ac.in} \\
  \And
  Balasubramanian Raman \\
  Dept. of Computer Science and Engineering\\
  Indian Institute of Technology\\
  Roorkee, India \\
  \texttt{bala@cs.iitr.ac.in} \\
}

\begin{document}
\maketitle

\begin{abstract}
Evaluation of students' performance for the completion of courses has been a major problem for both students and faculties during the work-from-home period in this COVID pandemic situation. To this end, this paper presents an in-depth analysis of deep learning and machine learning approaches for the formulation of an automated students' performance estimation system that works on partially available students' academic records. Our main contributions are (a) a large dataset with fifteen courses (shared publicly for academic research) (b) statistical analysis and ablations on the estimation problem for this dataset (c) predictive analysis through deep learning approaches and comparison with other arts and machine learning algorithms. Unlike previous approaches that rely on feature engineering or logical function deduction, our approach is fully data-driven and thus highly generic with better performance across different prediction tasks.
\end{abstract}

\keywords{Deep neural network \and COVID-19 \and Variational auto-encoder \and Educational institutions}

\section*{Introduction}\label{sec:introduction}
This work explores the problem of students' performance assessment under partial completion of their semester studies. This study is driven by the disruptions caused by COVID-19 in the education sector \cite{spinelli2020covid}. Course completion, organization of different exams, grades, admissions, student psychology, have been severely affected by this pandemic \cite{sahu2020closure,cao2020psychological}. Students worldwide are under tremendous stress due to uncertainty about their final grades. On the other hand, it has been very challenging for faculties to grade their students based on partial semester-completion with little to no means for assigning grades. We propose to solve this problem by leveraging deep learning for automatic prediction of students' grades. For this, different partial course-completion durations were experimented with.

Technology has played vital role in fighting against the COVID pandemic; however, educational plight has received little to no attention. This work is a humble attempt on leveraging Computational Intelligence (CI) methods for automatic students' performance assessment through predictive analysis. Such an automated system has numerous applications in the education sector. It is not only useful for the present pandemic situation but can also assist in admission decisions; non-completion; dropout and retention; profiling and prediction for student's feedback and guidance \cite{delnoij2020predicting,zawacki2019systematic}.

\begin{sidewaystable}
	\renewcommand{\arraystretch}{1.3}
	\caption{Literature review on approaches for academic performance assessment or prediction. Note that our work covers a large number of courses with much finer prediction range and better R\textsuperscript{2}-score.}
	\label{tab:lit}
	\centering
	\resizebox{.98\linewidth}{!}{
	\begin{tabular}{lcccccccccc}
		\toprule
		
		Author                                                    & Method                                                                                                                                    & Contribution                                                                                                                                                                                   & Dataset & \begin{tabular}[c]{@{}c@{}}\# of\\ courses\end{tabular} & \begin{tabular}[c]{@{}c@{}}Deep\\ Learning\end{tabular} & \begin{tabular}[c]{@{}c@{}}Machine\\ Learning\end{tabular} & \begin{tabular}[c]{@{}c@{}}Data\\ Analysis\end{tabular} & \begin{tabular}[c]{@{}c@{}}Prediction\\ range\end{tabular} & Metric                                                & Score                                      \\ \hline\hline
		\begin{tabular}[c]{@{}l@{}}Chen \textit{et al.}\\ (2014) \cite{chen2014training}\end{tabular}       & \begin{tabular}[c]{@{}c@{}}Multi-layer perceptron with\\ evolutionary algorithms, for\\ predictive analysis of\\ academic performance\end{tabular} & \begin{tabular}[c]{@{}c@{}}Comparison between\\ Cuckoo and Gravitational\\ search algorithms\end{tabular}                                                                                               & Small            & 3                                                                & \xmark                                                               & \cmark                                                                 & \xmark                                                               & \begin{tabular}[c]{@{}c@{}}Continuous\\ (0-10)\end{tabular}         & \begin{tabular}[c]{@{}c@{}}MAE,\\ RMSE,\\ MSE, R\end{tabular}  & \begin{tabular}[c]{@{}c@{}}0.72\\ (R\textsuperscript{2})\end{tabular} \\
		\begin{tabular}[c]{@{}l@{}}Livieris \textit{et al.}\\ (2016) \cite{livieris2016decision}\end{tabular}   & \begin{tabular}[c]{@{}c@{}}Multilayer perceptron,\\ SVM, machine\\ learning algorithms\end{tabular}                                                & \begin{tabular}[c]{@{}c@{}}A machine learning\\ algorithm interface (tool)\end{tabular}                                                                                                                 & Small            & 1                                                                & \xmark                                                               & \cmark                                                                 & \xmark                                                               & \begin{tabular}[c]{@{}c@{}}Categorical\\ (4 class)\end{tabular}     & Accuracy                                                       & 0.86                                                \\
		\begin{tabular}[c]{@{}l@{}}Li \textit{et al.}\\ (2016) \cite{li2016fuzzy}\end{tabular}         & \begin{tabular}[c]{@{}c@{}}Fuzzy clustering and\\ linear regression with\\ Random Forest and SVM\end{tabular}                                      & \begin{tabular}[c]{@{}c@{}}Fuzzy c-means\\ clustering implementation\end{tabular}                                                                                                                       & Large            & 3                                                                & \xmark                                                               & \cmark                                                                 & \xmark                                                               & \begin{tabular}[c]{@{}c@{}}Categorical\\ (20 class)\end{tabular}    & Accuracy                                                       & 0.79                                                \\
		\begin{tabular}[c]{@{}l@{}}Shehri \textit{et al.}\\ (2017) \cite{al2017student}\end{tabular}     & \begin{tabular}[c]{@{}c@{}}Academic performance\\ prediction using SVM\\ and k-NN\end{tabular}                                                     & \begin{tabular}[c]{@{}c@{}}Comparison between\\ machine learning\\ algorithms\end{tabular}                                                                                                              & Small            & 2                                                                & \xmark                                                               & \cmark                                                                 & \xmark                                                               & \begin{tabular}[c]{@{}c@{}}Categorical\\ (20 class)\end{tabular}    & \begin{tabular}[c]{@{}c@{}}MAE,\\ RMSE,\\ MSE, R\end{tabular}  & \begin{tabular}[c]{@{}c@{}}0.82\\ (R\textsuperscript{2})\end{tabular} \\
		\begin{tabular}[c]{@{}l@{}}Patil \textit{et al.}\\ (2019) \cite{patil2017effective}\end{tabular}      & \begin{tabular}[c]{@{}c@{}}LSTM based sequential\\ modeling of grades\end{tabular}                                                                 & \begin{tabular}[c]{@{}c@{}}Comparisong of deep\\ and non-deep methods\end{tabular}                                                                                                                      & Large            & 5                                                                & \cmark                                                              & \cmark                                                                 & \xmark                                                               & \begin{tabular}[c]{@{}c@{}}Categorical\\ (7 class)\end{tabular}     & \begin{tabular}[c]{@{}c@{}}RMSE,\\ Accuracy\end{tabular}       & 0.92                                                \\
		\begin{tabular}[c]{@{}l@{}}Harvey \textit{et al.}\\ (2019) \cite{harvey2019practical}\end{tabular}     & \begin{tabular}[c]{@{}c@{}}Machine learning for\\ feature analysis and\\ prediction in K-12\end{tabular}                                           & \begin{tabular}[c]{@{}c@{}}Predictive study of K-12\\ education dataset\end{tabular}                                                                                                                    & Large            & 1                                                                & \cmark                                                              & \xmark                                                                  & \cmark                                                              & \begin{tabular}[c]{@{}c@{}}Continuous\\ (0-100)\end{tabular}        & Accuracy                                                       & 0.71                                                \\
		\begin{tabular}[c]{@{}l@{}}Proposed\\ approach\end{tabular} & \begin{tabular}[c]{@{}c@{}}Deep learning methods\\ (LSTM, GRU, VAE)\\ with multiple machine\\ learning algorithms\end{tabular}                     & \begin{tabular}[c]{@{}c@{}}Integration of generative\\ and temporal deep\\ learning approaches with\\ machine learning algorithms\\ and statistical study of\\ the features of the dataset\end{tabular} & Large            & 15                                                               & \cmark                                                              & \cmark                                                                 & \cmark                                                              & \begin{tabular}[c]{@{}c@{}}Continuous\\ (0-100)\end{tabular}        & \begin{tabular}[c]{@{}c@{}}MAE,\\ RMSE,\\ MSE, R\textsuperscript{2}\end{tabular} & \begin{tabular}[c]{@{}c@{}}0.94\\ (R\textsuperscript{2})\end{tabular}             \\ \bottomrule
	\end{tabular}
	}
\end{sidewaystable}

Previously, operational-research based approaches have existed for solving this kind of problem. One example is the Duckworth/Lewis method for predicting adjusted target scores when the Cricket game is interrupted by rain  \cite{duckworth2004successful}. Duckworth \textit{et al.} studied thousands of Cricket match data and came out with a unique exponential function that tries to model the predicted score as a function of remaining resources. The problem of predictive performance assessment is challenging due to several reasons, and a Duckworth/Lewis approach seems naive since, due to the complexity of the problem, finding a model analytically is not possible. Estimation of marks requires modeling different kinds of correlations. Due to the uncertainty of human behavior, it is hard to conceptualize all relations needed for assessment, for example, missing dependencies such as previous scores in a subject, previous records of a student, etc. Although absolute predictions cannot be made, and there is always room for unforeseeable events, we have worked with a basic setting that considers the available marks. The proposed model tries to intelligently capture the latent correlations in students' performance and the complexity of subjects and other parameters.

Unprecedented pandemic circumstances and its adverse effect on education have propelled the relevance and need for a predictive assessment system based on available students' academic data. CI algorithms are a useful tool for addressing this. So far, very little work has happened on this problem. The scope of previous approaches is limited for experimenting through classical machine learning on smaller datasets with less number of courses. Most of them have focused on predicting categorical grades. There has been little to no study with deep learning. Previously, researchers have used evolutionary algorithms, multilayer perceptrons, fuzzy clustering, SVM, random forests for the predictive analysis of students' grades \cite{chen2014training,livieris2016decision,li2016fuzzy,al2017student,patil2017effective,harvey2019practical}. A summary of the related work on performance assessment methods is presented in Table \ref{tab:lit}. For a broader literature overview, we would suggest going through Hellas \textit{et al.} \cite{hellas2018predicting}.

Although there have been previous works with classical machine learning; however, through this paper, we present refreshing ideas into the field, leveraging deep learning. We show how deep learning can be used for students' academic score estimation. Unlike evolutionary algorithms employed by Chen \textit{et al.} \cite{chen2014training}, our emphasis is on gradient-based optimization algorithms since we get better generalizations, faster training, less computing resource requirement and these algorithms can virtually scale to any size of the dataset.  In our experiments, deep algorithms were found to perform better in comparison to machine learning approaches. Our dataset consists of fifteen courses. We have shown the integration of generative and temporal deep learning approaches with machine learning algorithms, and the statistical study of the features of the dataset. Our approach estimates final scores in a continuous range between (0, 100) rather than doing a grade based categorical estimation. The rest of the paper is organized as follows. The next section describes the dataset details and attributes. The proposed approach is presented after this in the section --- Approach. After this, the evaluation criteria is discussed. The results and conclusion are presented under the last two sections.

\section*{Dataset}\label{sec:dataset}
The data was collected at IIT Roorkee, India, for over 1000 students in 15 under-graduate and graduate level courses between 2006 to 2017. It has been anonymized to protect the identities of individual students. Students' academic performance is generally evaluated based on a set of individual parameters assessed across the comprehensive course of the evaluation. The dataset is called `IITR-APE dataset', and is made available at \href{https://hbachchas.github.io/data.html}{$ \texttt{https://hbachchas.github.io/data.html} $}.

Institutions use distinct sets of exams, like end-term evaluation and multiple in-course examinations, supplemented with an estimation of the class performance of the students. Some courses also involve laboratory work with hands-on experience over real-life applications. In the wake of COVID-19, almost every educational institution is facing the problem of deferred evaluation and staggering students' careers. This dataset is a humble effort towards the viability of recent developments in computational intelligence for the automatic assessment of students' performance.

Let $X$ be the set of all parameters used by the institutions for evaluation. The  evaluation system customarily composites a weighted average of the collection of all parameters in  $X$, which can be defined by the equation below:

\begin{eqnarray}
	Y &=& \sum_i W_i*X_i. \label{eq:1}
\end{eqnarray}

In equation \eqref{eq:1}, $W_i$ is the individual weight for each feature $X_i$, where $ X_i $ is the subset of feature-set $X$. The total score received by a student is represented by $Y$. The features used in our dataset include --- marks obtained in two tests $X_{T1}$ and $X_{T2}$, assessment based on students classroom performance $X_{CW}$, a mid-term evaluation $X_{MTE}$ and end-term evaluation $X_{ETE}$. Three datasets used for experimentation and evaluation consist of a subset of above mentioned features with specific weights for a given dataset as explained below.

Dataset $D_1$ consists of three basic features, which are --- two class-test based evaluations and one assessment based on the student's class performance. Hence, feature-set $X_{D1}$ can be defined as, $<X_{T1}, X_{T2}, X_{CW}>$. To establish an experimental setup for deferred evaluation for a given academic session, the midterm marks and end-term evaluation scores were dropped, and the remaining features were used to estimate the final score $Y_{D_1}$. This analysis helps us understand how features with small weightage, in score calculations, can effect the final score. 

Dataset $D_2$ consists of three features, viz., student's class performance, mid-term evaluation, and end-term evaluation. Hence, feature-set $X_{D2}$ can be defined as, $<X_{CW}, X_{MTE}, X_{ETE}>$. $Y_{D_2}$ represents a weighted sum of features in $X_{D2}$. Two distinct analyses were conducted over $X_{D2}$ to predict the final scores. First, using $X_{MTE}$ and $X_{CW}$, and later with $X_{ETE}$ and $X_{CW}$. Complementing $X_{MTE}$ with $X_{ETE}$, helped us analyze if deferred evaluation effects the total score estimation. On the other hand, utilizing $X_{ETE}$ helps to understand its effect on the overall score.

\section*{Approach}\label{sec:approach}
The datasets mentioned in the previous section on dataset description, include a set of different evaluation features for students' final score estimation. Different branches of deep learning and machine learning techniques were explored to evaluate the datasets. For temporal evaluation, Recurrent Neural Network (RNN) variants such as Long-Short Term Memory and Gated Recurrent Unit (GRU) were used along with discrete machine learning classifiers \cite{buckchash2020variational}. In cases where machine learning classifiers did not perform reasonably, Variational Bayes' encoding was employed for extensive feature extraction.

\subsection*{Long Short Term Memory (LSTM)}
Long Short Term Memory is a state-of-the-art neural architecture used for modeling complex long or short term temporal relations in sequence translation/recognition tasks. LSTM is incipiently an enhanced variant of the recurrent neural networks. LSTM structures are known for handling not only the hidden-state of an RNN, but also the cell-state of each recurring block. The mathematical description of a single LSTM cell is given by equation \eqref{eq:Input} through \eqref{eq:4}.

\begin{eqnarray}
	I_j &=& \sigma \left (\alpha_I [Z_{j-1},X_j ]+\beta_I \right). \label{eq:Input}\\
	F_j &=& \sigma \left (\alpha_F [Z_{j-1},X_j ]+\beta_F \right). \label{eq:forget}\\
	O_j &=& \sigma \left (\alpha_O [Z_{j-1},X_j ]+\beta_O \right). \label{eq:output}
\end{eqnarray}

The above equations represent the three different gates in the LSTM structure. $\sigma ()$ represents the sigmoid function in these equations. The symbol $\beta$ and $\alpha$ represent bias and weight matrices respectively, in each gate equation. $Z_{j-1}$ represents the hidden state of the LSTM cell from the previous time step, whereas $X_j$ represents the input to the LSTM cell at the current time step. Equation \eqref{eq:Input}, \eqref{eq:forget}, \eqref{eq:output} represent the input gate, forget gate and the output gate respectively. The input gate helps to capture the new information, to be stored in the cell state. The forget gate tells about the information, one needs to remove from the cell state. The output gate deduces the information that we need to emit as the final output from the LSTM cell. The outcomes of these three gates are used to find the LSTM's cell-state and hidden-state, as shown in the equations below:

\begin{eqnarray}
	\tilde{S}_j &=& \tanh \left( \alpha_S[Z_{j-1},X_j ] +\beta_S \right). \label{eq:2}\\
	S_j &=& F_j * S_{j-1} + I_j * \tilde{S}_j. \label{eq:3}\\
	Z_j &=& O_j * \tanh(S_j). \label{eq:4}
\end{eqnarray}

In the above equations, $S_j$ represents the cell-state memory for a given LSTM cell at timestamp $j$. $Z_j$ represents the hidden-state, at timestamp $j$, for the LSTM cell. 

LSTM has become a popular structure of choice, as it overcomes the problem of vanishing and exploding gradients in comparison to simple RNNs. In our analysis, the LSTM cells were joined with fully connected neural network layers to make predictions.

\subsection*{Gated Recurrent Unit (GRU)}\label{sec:GRU}
Gated Recurrent Unit is a modified version of recurrent neural networks. GRU resolves the problem of vanishing gradient, which dominates in standard RNN models. It is quite similar to the LSTM structure and sometimes even gives better performance. The working of GRU cells can be understood through the equations explained below:

\begin{eqnarray}
	H_j &=& \sigma \left( \alpha^H X_j+\gamma^H Z_{j-1} \right). \label{eq:update}
\end{eqnarray}

Equation \eqref{eq:update} represents the update gate inside a GRU cell. This gate helps to learn the amount of information the GRU cell needs to pass from the previous $(j-1)^{th}$ time step. $\alpha^H$ and $\gamma^H$ represent the weights of the new input, $X_j$, and the information from the previous $j-1$ time steps, $Z_{j-1}$, respectively.

\begin{eqnarray}
	R_j &=& \sigma \left( \alpha^R X_j+\gamma^R Z_{j-1} \right). \label{eq:Reset}
\end{eqnarray}

The above equation represents the reset gate. This gate helps the network to understand, how much of the information from the past $j-1$ time steps, need to be forgotten.

\begin{eqnarray}
	Z_j' &=& tanh \left( \alpha X_j+R_j \odot \gamma Z_{j-1} \right). \label{eq:current}
\end{eqnarray}

Equation \eqref{eq:current} helps in calculation of the current memory-state. This uses the input information --- $X_j$ and $Z_{j-1}$, along with $R_j$ from the reset gate, to store equivalent information from the past $j-1$ and the current time step.

\begin{eqnarray}
	Z_j &=& H_j\odot Z_{j-1} + (1-H_t) \odot Z_j'. \label{eq:final}
\end{eqnarray}

The above equation helps in the evaluation of the final hidden state information, $Z_j$, for the current time step to pass it on to a future time step.
In our experiments, the GRU units were integrated with two fully connected layers to make the grade prediction pipeline for students.

\subsection*{Variational Auto Encoder (VAE)}\label{sec:VAE}
Variational Auto Encoder is a common method used for feature extraction. The major difference between a simple Auto Encoder (AE) and a Variational Auto Encoder is that it learns the latent space variable $Z$ in the form of a prior distribution (usually a Gaussian).

In a VAE, the distribution of latent space is mapped with a presumed distribution. This distribution is learned in the format of mean $\mu$ and logarithmic variance $log\;\sigma$. To enforce this distribution to a prior distribution, we use $KL$ divergence loss which is defined as:

\begin{eqnarray}
	D_{KL}(N\left(\mu_i,\sigma_i\right),N\left(\mu_j,\sigma_j\right)) &=& log \frac{\sigma_j}{\sigma_i} + \frac{\sigma_i^2+(\mu_i-\mu_j)^2}{2\sigma_j^2}- \frac{1}{2}. \label{eq:KL}
\end{eqnarray}

The $KL$ divergence for two mapping Normal distributions is represented by equation \eqref{eq:KL}. Here, $N(\mu_j,\sigma_j)$ is our prior distribution and $N(\mu_i,\sigma_i)$ is the distribution upon which we want to enforce the prior distribution. The $\mu_i$ and $\sigma_i$ for the calculation are obtained from the latent space, $Z$, of the VAE. In our model, we have assumed the prior distribution to be $N(0,1)$ i.e. a Gaussian distribution with $\mu_j$ and $\sigma_j$ values of $0$ and $1$ respectively.

\begin{eqnarray}
	D_{KL}(N\left(\mu_i,\sigma_i\right),N\left(0,1\right)) &=& \frac{1}{2}(\sigma_i^2+\mu_i^2-1-2log\;\sigma_i). \label{eq:KLloss}
\end{eqnarray}

Equation \eqref{eq:KLloss} represents the $KL$ divergence loss for, $N(0,1)$, prior distribution used in our model.
The VAE model was trained on $X$ and the latent space $Z$ was extracted and further used to train the machine learning classifiers to make a final prediction of students' grades.

\subsection*{Machine Learning Classifier}\label{sec:ML}
The modeling of input data, $X$, to predict the students' grades, $Y$, was done using machine learning based classifiers. The classifiers used during our experiments are: Multi Layer Perceptron (MLP), Linear Regression (LR), Extra Tree Regressor (ET), Random Forest Regressor (RF), XGBoost Regreesor (XGB), and $ k $-Nearest Neighbour Regressor ($ k $NN). 

The variations in results were analyzed using 5-fold shuffle split cross-validation. For all our experiments, the dataset was randomly shuffled five times, and each time a test set was drawn to estimate the performance in terms of the metric results. The mean and standard deviation of these results are reported later under the section on results and discussion.

\begin{figure}[tb]
	\centering
	\includegraphics[width=.9\linewidth,keepaspectratio]{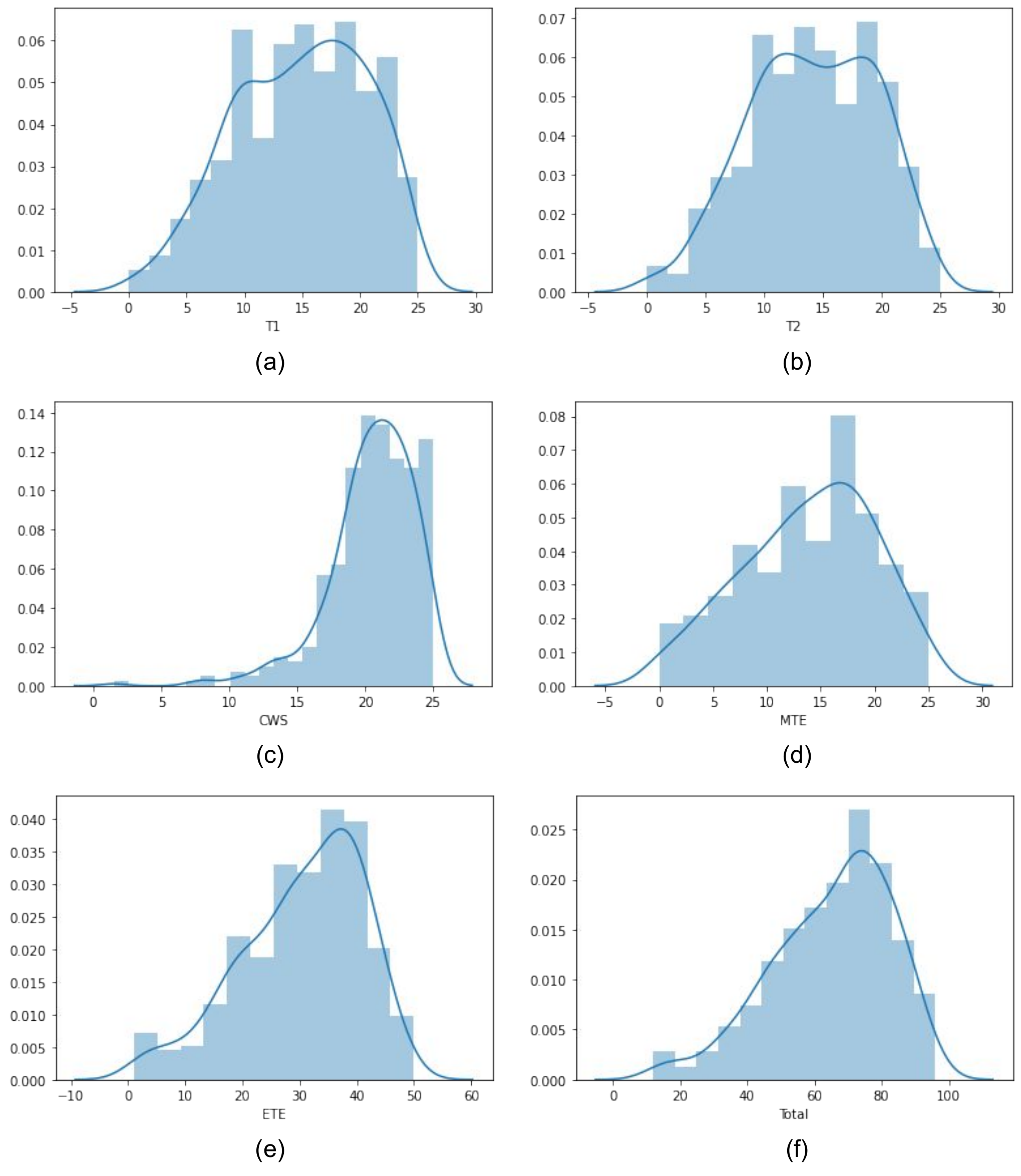}
	\caption{Distribution of data points for $D_1$ and $D_2$ datasets (a) test 1 ($X_{T1}$),(b) test 2 ($X_{T2}$), (c) class assessment ($X_{CW}$),(d) mid term evaluation marks ($X_{MTE}$),(e) end term evaluation marks ($X_{ETE}$),(f) total performance of student ($Y$).}
	\label{fig:1}
\end{figure}

\section*{Evaluation Criterion}\label{sec:evalcriter}
The models were evaluated using various evaluation criteria, which include R\textsuperscript{2}-score, Mean Absolute  Error (MAE), Mean Squared Error (MSE), and Root Mean Squared Error (RMSE). Calculation of all these metrics ensures a better evaluation of our regression models. The calculation involves true performance score $Y$, predicted performance score $\hat{Y}$, and the mean score $\Bar{Y}$.

\begin{eqnarray}
	\text{R}\textsuperscript{2}~Score &=& 1-\frac{\Sigma_k(Y_k-\hat{Y}_k)^2}{\Sigma_k(Y_k-\bar{Y})^2}.\\
	MAE &=& \frac{1}{\Sigma K}\Sigma_k\| Y_k-\hat{Y}_k\|.\\
	MSE &=& \frac{1}{\Sigma K}\Sigma_k( Y_k-\hat{Y}_k)^2.\\
	RMSE &=& \sqrt{MSE}.
\end{eqnarray}

R\textsuperscript{2}-score helps in deduction of the variation in the points along the regression line. These evaluation criteria metrics give an estimate of the effective predictions above the mean of the predicted label. MAE, MSE, RMSE help us to infer the error in predicting students' performance.

\begin{figure}[tb]
	\centering
	\includegraphics[width=.9\linewidth,keepaspectratio]{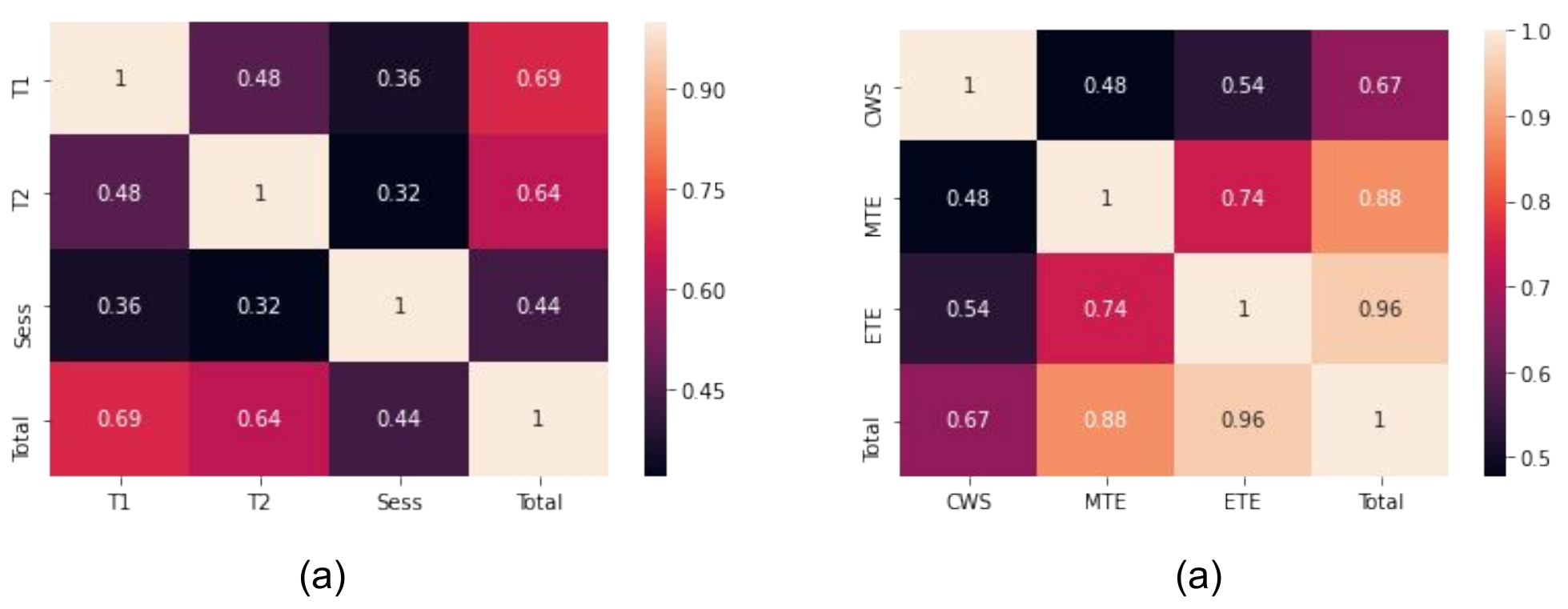}
	\caption{Correlation matrix for (a) dataset $D_1$ (b) dataset $D_2$.}
	\label{fig:2}
\end{figure}

\begin{table}[t]
	\renewcommand{\arraystretch}{1.3}
	\caption{Results for  dataset $D_1$ using features --- $X_{T1}$, $X_{T2}$, and $X_{CW}$; evaluation performed using --- R\textsuperscript{2}-score, MAE, MSE, RMSE.}
	\label{tab:D1}
	\centering
	\begin{tabular}{lcccc}
		\toprule
		& R\textsuperscript{2} Score            & MAE           & MSE           & RMSE           \\ \hline\hline
		VAE+MLP    & 0.561	& 7.601		&   124.679		& 11.129	       \\ \hline
		VAE+LR     & 0.585	&	7.191	&	116.077	&	10.744          \\ \hline
		VAE+ET       & 0.720& 5.943  &	77.709	&	8.781	     \\ \hline
		VAE+RF        & 0.720 &	6.264&	78.595 &	8.815         \\ \hline
		VAE+XGB & 0.714 & 6.211 & 	80.053	& 8.922         \\ \hline
		VAE+KNN & 0.584 & 	7.342 & 	115.858 & 	10.755 \\ \hline   
		LSTM & 0.587 & 138.230 & 11.757 & 7.356 \\ \hline 
		GRU & 0.672 & 77.214 & 8.787 & 6.431 
		\\ \bottomrule
	\end{tabular}
\end{table}

\section*{Results and Discussion}\label{sec:results and discussion}
The datasets mentioned previously, in dataset description section, can be analyzed based on  two methods: one, by using exploratory analysis, and other by using methods discussed in section --- Approach.

\subsection*{Exploratory Data Analysis}
Exploratory analysis of the datasets can be done using gradient maps, correlation matrix and distribution of data points. Fig. \ref{fig:1} shows the distribution of various features of the datasets. It is clearly visible that $X_{MTE}$ and $X_{ETE}$ have a similar distribution to the actual students' scores. However, the distribution of $X_{T1}$, $X_{T2}$, and $X_{CW}$ are quite different from the distribution of overall performance. Visual analysis of these plots suggests some correlation between the features and final prediction score.

\begin{figure}[tb]
	\centering
	\includegraphics[width=.97\linewidth,keepaspectratio]{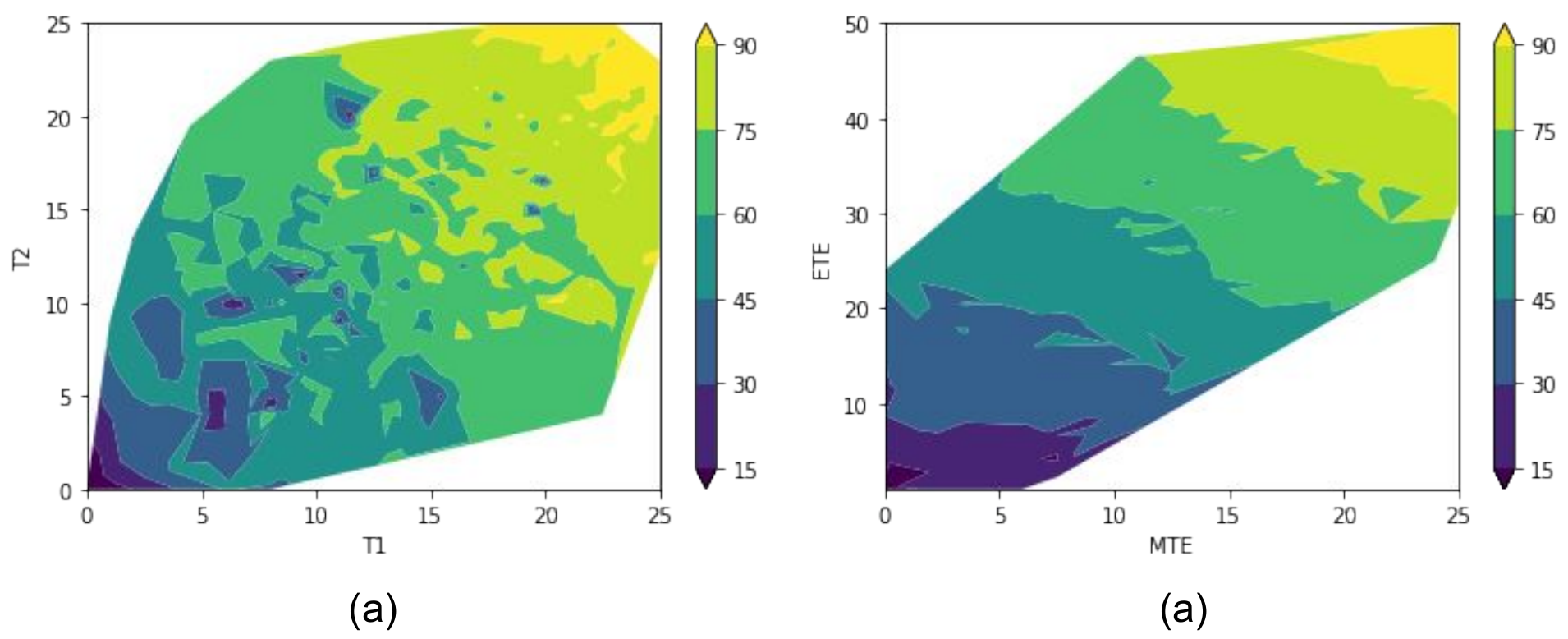}
	\caption{Gradient maps of students' performance for (a) $X_{T1}$ vs. $X_{T2}$ (B) $X_{ETE}$ Vs. $X_{MTE}$.}
	\label{fig:3}
\end{figure}

Moreover, the correlation matrix, in Fig. \ref{fig:2}, shows similar results with high correlation values of $0.88$ and $0.96$ for $X_{MTE}$ and $X_{ETE}$ respectively. Unlike that, $X_{T1}$ and $X_{T2}$ have a lower correlation score of $0.69$ and $0.64$ with respect to the total performance score of students'.
The correlations can further be verified with the gradient maps as shown in Fig. \ref{fig:3}. It clearly demonstrates that $X_{MTE}$ vs. $X_{ETE}$ has an evident trend of possible marks in comparison to $X_{T1}$ vs. $X_{T2}$.

\begin{table}[t]
	\renewcommand{\arraystretch}{1.3}
	\caption{Results for  dataset $D_2$ using features --- $X_{MTE}$ and $X_{CW}$; evaluation performed using --- R\textsuperscript{2}-score, MAE, MSE, RMSE.}
	\label{tab:D21}
	\centering
	\begin{tabular}{lcccc}
		\toprule
		& R\textsuperscript{2} Score            & MAE           & MSE           & RMSE           \\ \hline\hline
		MLP    & 0.867	&5.293&	43.595&	6.590	       \\ \hline
		LR     & 0.866 & 5.325 & 44.142	& 6.633 \\ \hline
		ET       & 0.796 & 	6.352 &	67.329 &	8.173     \\ \hline
		RF        &0.823 &	5.975 & 	58.386 &	7.621         \\ \hline
		XGB & 0.808 &  6.139&	63.157&	7.918 \\ \hline
		KNN & 0.842&	5.727&	52.204&	7.203 \\ \hline   
		LSTM & 0.845 & 5.180& 43.748 & 6.614 
		\\ \hline 
		GRU &  0.850& 4.954  & 42.362 & 6.508    \\ \bottomrule
	\end{tabular}
\end{table}

\subsection*{Prediction Results}
The results of predictive modeling using various approaches described under section --- Approach --- are given in Table \ref{tab:D1}, \ref{tab:D21}, and \ref{tab:D22}.
For dataset $D_1$, the results using VAE with machine learning classifiers along with GRU and LSTM are reported. It is evident that minimum error was acquired by VAE in conjunction with Extra Tree Regressor classifier with R\textsuperscript{2}-score, MAE, MSE, and RMSE of 0.720, 5.943, 77.709, and 8.781 respectively. The results show sufficient utility of $X_{T1}$ and $X_{T2}$ for students' performance prediction.

For dataset $D_2$, the results using two different experiments are shown. In Table \ref{tab:D21}, using $X_{MTE}$ and $X_{CW}$, it is evident that minimum error was acquired by VAE with Multi Layer Perceptron Regressor with R\textsuperscript{2}-score, MAE, MSE, and RMSE values of 0.947, 3.249, 14.885, and 3.858  respectively. The results show that due to substantial weightage and maximum efforts given by the students, $X_{ETE}$ has the highest impact on the evaluation of a student's collective performance.

\begin{table}[t]
	\renewcommand{\arraystretch}{1.3}
	\caption{Results for  dataset $D_2$ using features --- $X_{ETE}$ and $X_{CW}$; evaluation performed using --- R\textsuperscript{2}-score, MAE, MSE, RMSE.}
	\label{tab:D22}
	\centering
	\begin{tabular}{lcccc}
		\toprule
		& R\textsuperscript{2} Score            & MAE           & MSE           & RMSE           \\ \hline\hline
		MLP    & 0.943	& 3.495 & 17.171 & 4.141    \\ \hline
		LR     & 0.947 &	3.385	&17.279&	4.154\\ \hline
		ET       & 0.918&	4.104&	26.646&	5.144    \\ \hline
		RF        &0.928&	3.901&	23.430&	4.829       \\ \hline
		XGB & 0.929 &	3.881&	22.976&	4.788 \\ \hline
		KNN & 0.933	&3.798&	21.706&	4.649	 \\ \hline   
		LSTM & 0.946& 3.263 & 15.078 & 3.883  \\ \hline 
		GRU &  0.947& 3.249  & 14.885 & 3.858    \\ \bottomrule
	\end{tabular}
\end{table}

\begin{table}[t]
	\renewcommand{\arraystretch}{1.3}
	\caption{Comparison of the proposed approach with other related works. R\textsuperscript{2} and accuracy metrics are used. It can be noted that the proposed method has a significant improvement in R\textsuperscript{2} score.}
	\label{tab:result_comp}
	\centering
	\begin{tabular}{lcc}
		\toprule
		Method                                                & Metric & Score \\ \hline\hline
		Chen \textit{et al.} \cite{chen2014training}, 2014   & R\textsuperscript{2}              & 0.72           \\
		Shehri \textit{et al.} \cite{al2017student}, 2017 & R\textsuperscript{2}              & 0.82           \\
		Olive \textit{et al.} \cite{olive2019quest}, 2019  & Acc.            & 0.77           \\
		Proposed approach                                                & R\textsuperscript{2}              & \textbf{0.94}           \\ \bottomrule
	\end{tabular}
\end{table}

In Table \ref{tab:D22}, for $X_{ETE}$ and $X_{CW}$, it is evident that minimum error was acquired by GRU with a R\textsuperscript{2}-score, MAE, MSE, and RMSE of 0.867, 5.293, 43.595, and 6.590 respectively. The results establish the utility of $X_{MTE}$ and $X_{CW}$ for predicting students' performance in case of a deferred evaluation of an ongoing academic session. Of all the models tested by us, the best results were obtained by LSTM and GRU methods, as shown in Table \ref{tab:result_comp}. Our models attained fairly large R\textsuperscript{2}-score, in comparison to other previous works, which shows how good deep learning techniques are for evaluation and prediction purposes. It should be noted that there is significant scope of increment in the size of our dataset, and it should only add to further improvement of the proposed deep learning approaches.

\section*{Conclusion}\label{sec:conclusion}
This study features the role of CI in alleviation of challenges and impact of COVID pandemic on education. Application of deep learning methods for academic performance estimation is shown. State of the current arts is explained with conclusive related work. For the purpose of evaluation and benchmarking, an anonymized students' academic performance dataset, called IITR-APE, was created and will be released in the public domain. The promising performance of the proposed approach explains the suitability of modern-day CI methods for modeling students' academic patterns. However, we feel that availability of larger datasets would further allow the system to be more accurate. For this, we are building a larger pubic version of the current dataset. It was observed through the performance stats and gradient maps that better prediction happens when all components of final grade have equal weightage. It is thus suggested to have a continual evaluation strategy, i.e., there should be many equally weighted tests or assignments, conducted regularly with short frequency, rather than conducting two or three high weightage exams.

This work is a humble attempt in the direction of enabling CI based methods for handling adverse effects of COVID like pandemics on education. It is expected that this study may pave the path for future research in the direction of CI enabled predictive assessment of students' marks.

\section*{Acknowledgments}
This work was supported by the Ministry of Electronics and Information Technology, Gov. of India, under Visvesvaraya PhD Fellowship. The dataset was contributed by --- Indian Institute of Technology, Roorkee.


%
%

%

%


\bibliographystyle{unsrt}  
\bibliography{references}  

%
%
%
%

\end{document}